\documentclass[11pt,a4paper]{article}
\usepackage{times,latexsym}
\usepackage{url}
\usepackage[T1]{fontenc}
\usepackage[acceptedWithA]{tacl2021v1}

\usepackage[utf8]{inputenc}
\usepackage{microtype}
\usepackage{amsmath}
\usepackage[linguistics]{forest}
\usepackage{caption,subcaption}
\usepackage{multirow}
\usepackage{amssymb}
\usepackage{graphicx}
\usepackage{tabularx}
\usepackage{booktabs}
\usepackage{tikz}
\usepackage[autolanguage]{numprint}
\usepackage{algorithm}
\usepackage{algorithmicx}
\usepackage{algpseudocode}
\usepackage{fdsymbol}

\algrenewcommand\algorithmicrequire{\textbf{Input:}}
\algrenewcommand\algorithmicensure{\textbf{Output:}}

\graphicspath{ {./tacl_figures/} }

\newenvironment{itemizesquish}{\begin{list}{\labelitemi}{\setlength{\itemsep}{0em}\setlength{\labelwidth}{0.5em}\setlength{\leftmargin}{\labelwidth}\addtolength{\leftmargin}{\labelsep}}}{\end{list}}

\title{Transformer Grammars: Augmenting Transformer Language Models with Syntactic Inductive Biases at Scale}

\author{Laurent Sartran$^1$ \, Samuel Barrett$^2$ \, Adhiguna Kuncoro$^{1,2}$\\
\textbf{ Milo\v{s} Stanojevi\'c$^1$ \, Phil Blunsom$^2$ \, Chris Dyer$^1$}\\
$^1$DeepMind, $^2$University of Oxford\\
\texttt{\{lsartran,akuncoro,stanojevic,cdyer\}@deepmind.com}\\
\texttt{samuelbarrett1234@btinternet.com, phil.blunsom@cs.ox.ac.uk}\\
}
  
\newcommand{\figContentActionsExample}{
\begin{tabular}{l | l | l | l | l} 
  $i$   & Input $a'_i$ &  Type &    Attn. op. & Label \\
  \hline
  0     & \texttt{<s>}  & \texttt{ONT}   & \textsc{stack} & (S         \\
  1     & (S            & \texttt{ONT}   & \textsc{stack} & (NP        \\
  2     & (NP           & \texttt{ONT}   & \textsc{stack} & the        \\
  3     & the           & \texttt{T}     & \textsc{stack} & blue       \\
  4     & blue          & \texttt{T}     & \textsc{stack} & bird       \\
  5     & bird          & \texttt{T}     & \textsc{stack} & NP)        \\
  6     & NP)           & \texttt{CNT1}  & \textsc{compose} & --         \\
  7     & NP)           & \texttt{CNT2}  & \textsc{stack} & (VP        \\
  8     & (VP           & \texttt{ONT}   & \textsc{stack} & sings      \\
  9     & sings         & \texttt{T}     & \textsc{stack} & VP)        \\
  10    & VP)           & \texttt{CNT1}  & \textsc{compose} & --         \\
  11    & VP)           & \texttt{CNT2}  & \textsc{stack} & S)         \\
  12    & S)            & \texttt{CNT1}  & \textsc{compose} & --        \\
  13    & S)            & \texttt{CNT2}  & \textsc{stack} & --      \\
  \hline
\end{tabular}
}

\newcommand{\figureExampleTransitionsAndAttentionMatrix}{
  \begin{figure*}
      \centering
      \begin{tabular}{cc}
      {
      \subcaptionbox{Example of a (transformed) sequence with its corresponding token types, type of attention operations, and labels. No prediction is done for positions 6, 10, 12 (where \textsc{compose} is performed), nor for position 13 as no end-of-sequence token is required to model linearized trees.}{
        \scalebox{1.0}{\mbox{\figContentActionsExample{}}}
        \label{table:sequence}
      }} & %${}\qquad{}$ &
      \subcaptionbox{Attention mask with \textsc{stack}/\textsc{compose} attention. \textsc{stack} is represented in blue, whereas \textsc{compose} is denoted in orange.}{
        \mbox{\includegraphics[scale=1.0]{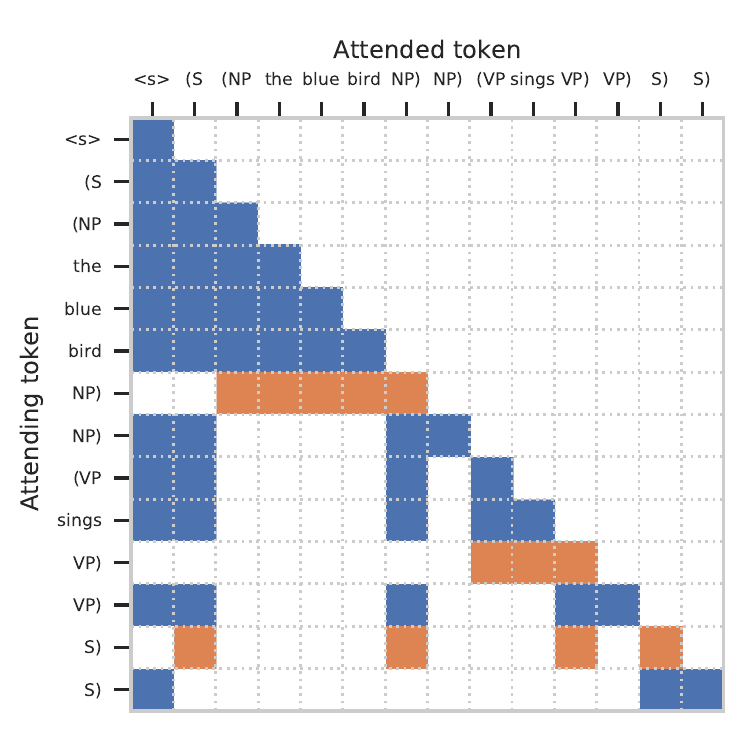}}
        \label{mask}
      }
      \end{tabular}
     \caption{Processing of an example sentence: {(S (NP the blue bird NP) (VP sings VP) S)}}

      \label{fig:processing}
  \end{figure*}
}

\begin{document}
\maketitle
\begin{abstract}

We introduce Transformer Grammars (TGs), a novel class of Transformer language models that combine (i) the expressive power, scalability, and strong performance of Transformers and (ii) recursive syntactic compositions, which here are implemented through a special attention mask and deterministic transformation of the linearized tree. We find that TGs outperform various strong baselines on sentence-level language modeling perplexity, as well as on multiple syntax-sensitive language modeling evaluation metrics. Additionally, we find that the recursive syntactic composition bottleneck which represents each sentence as a single vector harms perplexity on document-level language modeling, providing evidence that a different kind of memory mechanism---one that is independent of composed syntactic representations---plays an important role in current successful models of long text. 
\end{abstract}

\section{Introduction}

Transformer language models (LMs) that are trained on vast amounts of data have achieved remarkable success at various NLP benchmarks \citep[\emph{inter alia}]{peters_2018,devlin_2019,brown_2020}. Intriguingly, this success is achieved by models that lack an explicit modeling of hierarchical syntactic structures, which were hypothesized by decades of linguistic research to be necessary for good generalization \citep{chomsky_1957,structures:not:strings}. This naturally leaves a question: To what extent can we \emph{further improve} the performance of Transformer LMs, through an inductive bias that encourages the model to explain the data by the means of recursive syntactic compositions? Although the benefits of modeling recursive syntax have been shown at the small data and model scales, such as in the case of recurrent neural network grammars \citep[\emph{inter alia}]{dyer-etal-2016-recurrent,futrell_2019,kim_2019,Hu:et-al:2020,noji-oseki-2021-effective}, it remains an open question whether---and to what extent---a similar design principle is still beneficial for Transformer LMs at larger scales.

In this paper, we aim to answer these questions by introducing \textbf{Transformer Grammars} (TGs)---a novel class of Transformer language models that combine: (i) the expressive power, scalability, and strong performance of Transformer-XL \citep{dai-etal-2019-transformer}; (ii) joint modeling of surface strings $\boldsymbol{x}$ and their corresponding phrase-structure trees $\boldsymbol{y}$, \emph{i.e.,} $p(\boldsymbol{x}, \boldsymbol{y})$; and (iii) an inductive bias that constrains the model to explain the data through built-in recursive syntactic composition operations. By implementing these recursive compositions through a novel modification of the Transformer-XL attention mask, TGs retain the computational efficiency of standard Transformer-XLs, enabling them to avoid the limitations of LSTM-based recurrent neural network grammars~\citep[RNNGs]{dyer-etal-2016-recurrent}, which have been proven difficult to scale~\citep{kuncoro_19,noji-oseki-2021-effective}.

TGs are related to the recent work of \citet{qian-etal-2021-structural} that similarly aims to augment generative Transformer language models with a stronger modeling of syntactic structures, albeit with two key differences. First, whereas \citet{qian-etal-2021-structural} used syntactic structure to restrict the behavior of a subset of attention heads \citep{strubell-etal-2018-linguistically,astudillo-etal-2020-transition}, TGs incorporate a stronger form of syntactic inductive bias by using recursive syntactic compositions to create an explicit composed representation for each constituent, in a similar fashion as RNNGs. Hence, our approach sheds light into whether, and to what extent, the recursive syntactic composition hypothesis---which has been shown to be valuable at the small data and model scale in the case of RNNGs---continues to offer additional benefits, beyond specializing a subset of attention heads for syntax. Second, in contrast to prior work, which has been limited to modeling sentences independently of document context, this work explores whether syntactic composition also benefits \emph{document-level} language modeling.

We evaluate Transformer Grammars against baseline models on three metrics: (i) perplexity, (ii) syntactic generalization, and (iii) parse reranking, and on two training datasets: (i) the small-scale Penn Treebank \citep[PTB]{marcus93} and (ii) the medium-scale BLLIP-\textsc{lg} \citep{charniak2000bllip} datasets, with $\sim$1M and $\sim$40M words, respectively. We find that:
\begin{itemizesquish}
\item Transformer Grammars (\textbf{TGs}) achieve better (i) single-sentence language modeling perplexity, (ii) syntactic generalization, and (iii) parse reranking performance than RNNGs---all the while being much faster to train than the batched RNNG of \citet{noji-oseki-2021-effective}.
\item In single-sentence language modeling perplexity, the terminal-only Transformer XL baseline~(\textbf{TXL (terminals)}) is outperformed by both TGs and a Transformer XL that predicts sentences as joint sequences of terminals and tree-building nonterminal symbols~(\textbf{TXL (trees)}) but \emph{without} the TG's attention restrictions~\citep{choe-charniak-2016-parsing,qian-etal-2021-structural}.
\item Although modeling structure improves perplexity compared to terminal-only models, the TXL~(trees) model slightly outperforms the more biased TG models. Using a regression analysis, we show that while TG's recursive syntactic compositions benefit syntactic generalization, their implementation in terms of attention restriction interferes with Transformers' lexical copying ability, which turns out to play a role in obtaining low perplexities. This result indicates a partial dissociation between perplexity and syntactic generalization, both of which are important metrics for assessing LM success.
\item On the benchmark of \citet{Hu:et-al:2020} that is a carefully controlled test of syntactic generalization ability, TGs substantially outperform the syntax-free TXL (terminals) baseline, as well as the much stronger TXL (trees) model. Perhaps even more remarkably, TGs outperform the GPT-2-small \citep{radford_2019}, Gopher \citep{rae_2021}, and Chinchilla \citep{chinchilla} models, which are between $250\times$ and $1,000\times$ larger than the TG model, trained on vastly more data, and arguably represent the most sophisticated LMs in existence.
\item When modeling full documents and evaluating perplexity, we again find that the TXL (trees) model outperforms the terminal-only TXL (terminals) baseline. However, TGs substantially underperform both the TXL (terminals) and TXL (trees) models. The failure of TGs, which represent prior-sentence context purely in terms of a composed syntactic representation, suggests that a different memory mechanism---that works in part independently of syntactic structures---may play a role in the processing of long-form text. 
\end{itemizesquish}

All in all, our findings show that---under comparable experimental conditions---LMs \emph{with} notions of syntactic structures (both TXL (trees) \& TG) outperform those \emph{without} on multiple evaluation metrics. We further demonstrate that encouraging the model to explain the data through the means of recursive syntactic compositions---as is the case for TGs---is a valuable inductive bias for achieving an \emph{even stronger} human-like syntactic competence, outperforming prior work that also incorporates syntactic biases, albeit without recursive compositions \citep{qian-etal-2021-structural}, in addition to some of the largest non-syntactic LMs to date. Lastly, our findings motivate the development of \emph{scalable} LMs---that nevertheless incorporate stronger notions of syntactic structures---as a promising (albeit relatively under-explored) area of NLP research.

\section{Model}

\begin{figure}[t]
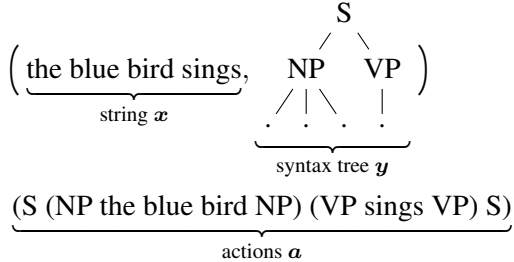

\centering
\tikzstyle{level 1}=[level distance=0.75cm, sibling distance=1cm]
\tikzstyle{level 2}=[level distance=0.75cm, sibling distance=0.5cm]

\begin{align*}
& \Big(\underbrace{\text{the blue bird sings}}_{\text{string }\boldsymbol{x}},
\underbrace{\tikz[baseline=(NP.base)]{
]
\node{S}
child {node (NP) {NP}
    child {node {.}}
    child {node {.}}
    child {node {.}}
}
child {node {VP} 
    child {node {.}}
};
}}_{\text{syntax tree }\boldsymbol{y}}\Big) \\
& \underbrace{\text{(S (NP the blue bird NP) (VP sings VP) S)}}_{\text{actions }\boldsymbol{a}}
\end{align*}
\caption{An example that represents a pair of string $\boldsymbol{x}$ and its phrase-structure tree $\boldsymbol{y}$, which are then represented as a sequence of actions that construct ($\boldsymbol{x}$, $\boldsymbol{y}$) in a top-down, left-to-right fashion \cite{dyer-etal-2016-recurrent,choe-charniak-2016-parsing}.}
\label{choe_charniak}
\end{figure}

TGs are syntactic language models: they jointly model the probability of syntactic phrase-structure trees $\boldsymbol{y}$ and strings of words $\boldsymbol{x}$, using the predicted structures to determine the structure of the computations of model states. Following a line of recent work in parsing and syntactic language models \citep{vinyals:2015,dyer-etal-2016-recurrent,choe-charniak-2016-parsing}, the generation problem is decomposed into modeling a sequence of \emph{actions} that construct ($\boldsymbol{x}$, $\boldsymbol{y}$) in a top-down, left-to-right fashion, by interleaving nonterminal nodes and their children, as shown in Figure~\ref{choe_charniak}. The linearized representation of ($\boldsymbol{x}$, $\boldsymbol{y}$) consists of three types of actions: (i)~opening nonterminals (action type \texttt{ONT}), marking the opening of a new constituent; (ii)~generating terminal symbols/leaf nodes (\emph{i.e.,} words or subword tokens), henceforth denoted as \texttt{T}; and (iii)~closing the most recent open constituent/incomplete nonterminal symbol, henceforth denoted as \texttt{CNT}.

Let $\boldsymbol{a} = (a_0, a_1, \ldots, a_{T-1})$ be a sequence of actions (of length $T$) that generates ($\boldsymbol{x}$, $\boldsymbol{y}$), where each action is part of the action vocabulary $\mathcal{V}$. TGs define a probability distribution over $\boldsymbol{a}$ through a left-to-right factorization, \emph{i.e.,} $p(\boldsymbol{x}, \boldsymbol{y}) = p(\boldsymbol{a}) = \prod_i p(a_i \mid \boldsymbol{a}_{< i})$.

\subsection{Recursive syntactic composition via attention}\label{sec:recursive_composition}

In Transformer language models, when generating $a_i$ conditionally on $\boldsymbol{a}_{<i}$, attention is the only mechanism by which information from other positions $j < i$ is incorporated. The rules governing this information flow---\emph{i.e.,} which positions can attend to which other positions---are defined by the \emph{attention mask}. We design TGs to use recursive syntactic compositions, which have been shown to lead to better generalisation in the LSTM-based RNNG model, and we implement them through the Transformer attention mechanism.

In TGs, the action sequence is generated from left-to-right, and each symbol $a_i$ can be thought of as updating a stack of indices. When the current $a_i$ is a closing nonterminal (\emph{i.e.}, a constituent has just ended) its index $i$ will be represented by a single-vector-sized \emph{composed} representation obtained by attending to the child positions of the currently ending constituent. Subsequent positions ($>i$) may attend to this composed position, but they may not attend directly to the constituent positions, and this restriction imposes a syntactic bottleneck, since everything inside the constituent that influences subsequent predictions must become part of the composed representation. This bottleneck encourages the model to learn informative representations of composed phrases and is inspired by a similar design principle as RNNGs and other tree-structured architectures. In the stack, the restriction is instantiated by popping the indices for the child nodes and then pushing the index of the composed constituent. We refer to this process as \textsc{compose} attention. 

In addition to \textsc{compose} attention, at each position $i$, we apply \textsc{stack} attention, where $i$ is pushed onto the stack attention, and attention is restricted to positions on the stack. Both \textsc{stack} and \textsc{compose} attention use the same parameters and attention heads---what distinguishes them is only the rule for computing the set of positions that the model can attend to. Importantly, as we need to (i) perform \emph{both} \textsc{compose} and \textsc{stack} for a closing nonterminal (\emph{e.g.,} to first compute a composed representation based on its parts/children, and then add the composed representation onto the stack), while (ii) performing exactly one attention operation per token, we transform the original sequence $\boldsymbol{a}$ by duplicating all closing nonterminals. This yields a sequence $\boldsymbol{a}'$ of length $T'$, \emph{e.g.,} \texttt{(S (NP the blue bird NP) NP) (VP sings VP) VP) S) S)}. The first closing nonterminal of each pair is given the type \texttt{CNT1}, and implements \textsc{compose}, whereas the second is given the type \texttt{CNT2}, and implements \textsc{stack}. To keep the number of prediction events (\emph{i.e.,} the number of times a probability distribution is emitted by the model) constant, no final prediction is made for \textsc{compose} positions (see Figure~\ref{fig:processing}).

\figureExampleTransitionsAndAttentionMatrix{}

The exact procedure for \textsc{stack}/\textsc{compose} is described in Algorithm~\ref{stack/compose attention}. The positions that may be attended are represented as a binary attention mask $\mathbf{A} \in \mathbb{R}^{T' \times T'}$ (see Figure~\ref{fig:processing}), such that $A_{ij} = 1$ \emph{iff} the position $j$ may be attended from $i$, and 0 otherwise. Note that the computation of the attention mask is causal, \emph{i.e.,} no information from positions $j > i$ is used to compute the positions that can be attended from $i$.

\paragraph{Relative positional encoding.}\label{sec:relpos} In Transformer-XL, the positional information presented to the model is based on the difference between the attending position $i$ and the attended position $j$, \emph{i.e.}, $i - j$. This distance does not reflect nor use the topology of the tree. We thus generalize how relative positions are provided to the attention mechanism such that any matrix $\mathbf{R} \in \mathbb{Z}^{T' \times T'}$ can be used, where $R_{ij}$ is the relative position between $i$ and $j$. For TGs, we define $R_{ij} = \delta(i) - \delta(j)$, where $\delta(i)$ is the depth of the $i$-th token in the tree.
% Note that two tokens at the same depth such that one can attend to the other are siblings in the tree.
Note that the relative distance $R_{ij}$ will only be computed if $A_{ij} = 1$ (\emph{i.e.,} $j$ may be attended from $i$). For instance, for the action sequence in Figure~\ref{choe_charniak}, the relative distance between (the positions corresponding to) the words \texttt{sings} and \texttt{bird} is never computed, but it will be computed between \texttt{sings} and its sibling NP covering \texttt{the blue bird}.

\begin{algorithm}
\begin{small}
\caption{\textsc{stack}/\textsc{compose} attention}
\label{stack/compose attention}
\begin{algorithmic}[1]

\Require $\boldsymbol{a}'$ sequence of tokens
\Ensure $\mathbf{A} \in \mathbb{R}^{T' \times T'}$ attention mask
\State $S \gets []$\Comment{Empty stack}
\State $\mathbf{A} \gets 0$
\For{$i \gets 0$ to $T'$}
    \If{$\textrm{type}(\boldsymbol{a}'[i]) = \texttt{CNT1}$}
        \Comment{\textsc{compose}}
        \State $j \gets i$
        \While{$\textrm{type}(\boldsymbol{a}'[j]) \neq \texttt{ONT}$}
            \State $A_{ij} \gets 1$
            \State $j \gets S.pop()$
        \EndWhile
        \State $A_{ij} \gets 1$
        \State $S.push(i)$
    \Else
        \Comment{\textsc{stack}}
        \If{$\textrm{type}(\boldsymbol{a'}[i]) \neq \texttt{CNT2}$}
            \State $S.push(i)$
        \EndIf
        \For{$j \in S$}
            \State $A_{ij} \gets 1$
        \EndFor
    \EndIf
\EndFor
\State \textbf{return} $\mathbf{A}$\Comment{Attention mask}
\end{algorithmic}\end{small}
\end{algorithm}

\subsection{Segmentation and recurrence}

In the same manner as Transformer-XL, Transformer Grammars are recurrent neural networks that can process arbitrarily long sequences\footnote{This desirable property of Transformer-XL is the main justification for our using it as baseline and starting point.} as consecutive segments that contain a fixed number of tokens $L$, maintaining and updating a memory of temporal dimension $M$ from one segment to the next. With $0~\le~\tau~\le~\lceil~\frac{T'}{L}~\rceil$, $\boldsymbol{a}'_\tau = \left(a_{\tau L}, a_{\tau L+1}, \ldots, a_{\tau \left(L+1\right)-1}\right)$ is the $\tau+1$-th segment. Token embeddings are obtained from an embedding matrix $\mathbf{E} \in \mathbb{R}^{|\mathcal{V}| \times d}$ to form a sequence of $L$ vectors in $\mathbb{R}^d$: $\mathbf{h}^{(0)}_\tau~=~\left(h^{(0)}_{\tau L}, \ldots, h^{(0)}_{\tau \left(L+1\right)-1} \right)$.

The core of the model is composed of $K$ stacked recurrent layers, \emph{i.e.}, for $1 \le k \le K$: $$ \mathbf{h}^{(k)}_\tau, \mathbf{m}^{(k)}_{\tau+1} = \mathrm{Layer}^{(k)}(\mathbf{h}^{(k-1)}_\tau, \mathbf{m}^{(k)}_{\tau}, \mathbf{A}_\tau, \mathbf{R}_\tau)$$ where for each segment $\tau$: \begin{itemize}
\item $\mathbf{h}^{(k)}_\tau~\in~\mathbb{R}^{L \times d}$ is the sequence of hidden states, which forms the input for layer $k+1$,
\item $\mathbf{m}^{(k)}_{\tau}~\in~\mathbb{R}^{M \times d}$ is the memory,
\item $\mathbf{A}_\tau~\in~\mathbb{R}^{L \times (M+L)}$ is the attention mask from the current segment to the current segment and the memory,
\item and $\mathbf{R}_\tau~\in~\mathbb{Z}^{L \times (M+L)}$ is the corresponding relative positions matrix.
\end{itemize} All layers receive the same attention mask and relative positions matrix. Each layer $k$ is composed of a multi-head self-attention ($\mathrm{SelfAttn}$) sub-layer and a position-wise feed-forward network ($\mathrm{FFN}$) sub-layer (with residual connections followed by layer normalization---omitted for clarity), as well as an update to the memory for the next segment: \begin{align*}
\mathbf{h}^{(k-\frac{1}{2})}_\tau &= \mathrm{SelfAttn}_k(\mathbf{h}^{(k-1)}_\tau, \mathbf{m}^{(k)}_{\tau}, \mathbf{A}_\tau, \mathbf{R}_\tau) \\
\mathbf{h}^{(k)}_\tau &= \mathrm{FFN}_k(\mathbf{h}^{(k-\frac{1}{2})}_\tau) \\
\mathbf{m}^{(k)}_{\tau+1} &= \mathrm{MemoryUpdate}(\mathbf{h}^{(k-1)}_\tau, \mathbf{m}^{(k)}_{\tau})
\end{align*}
The output of the last layer, $\mathbf{h}^{(K)}_\tau$, is multiplied by the transpose of the embedding matrix $\mathbf{E}^T$ to get the unnormalized next-token log probabilities.

\paragraph{Self-attention.} Using the notation of~\citet{dai-etal-2019-transformer}, let $\mathbf{W}_q$, $\mathbf{W}_{k,E}$, $\mathbf{W}_{k,R}$, $\mathbf{W}_v$, and $u$ and $v$ be the trainable model parameters. Let $\left[ \cdot, \cdot \right]$ denote a concatenation operation along the time dimension. For a single head, we have: \begin{align*}
\mathbf{q} &= \mathbf{h} \mathbf{W}_q & 
\mathbf{k} &= \left[ \mathbf{m}, \mathbf{h} \right] \mathbf{W}_{k,E} &
\mathbf{v} &= \left[ \mathbf{m}, \mathbf{h} \right] \mathbf{W}_{v}.
\end{align*}
The attention score for an attending position $i$ and an attended position $j$ is \begin{equation*}
    s_{ij}= (\mathbf{q}_i + u)^T \mathbf{k}_j + (\mathbf{q}_i + v)^T \mathbf{r}_{ij},
\end{equation*}
where $\mathbf{r}_{ij} \in \mathbb{R}^{d}$ is an embedding of the integer relative position $R_{ij}$ (row from $\mathbf{W}_{k,R}$). Much like in Transformer-XL, the second term can be computed efficiently as the relative positions take values within a small interval $\left[R_\textrm{min}, R_\textrm{max}\right]$.

The mask $\mathbf{A}$ (\S\ref{sec:recursive_composition}) is applied element-wise on the scores, which sets masked entries to $-\infty$. The normalized attention weights are then obtained by applying a $\mathrm{softmax}$ activation function to the scores; the final attention outputs are the product of the attention weights and the values. In practice, we use multiple heads---the outputs of each are concatenated and passed onto a linear transformation.

\paragraph{Memory update.} In Transformer-XLs, the memory is updated by shifting the current input into it. Here we take advantage of the fact that positions within a subtree that have been \textsc{compose}d are never attended to in the future, and \emph{a fortiori} in the following segments. Hence, only positions that may be attended need to be added or kept in the memory. This requires careful book-keeping of which position in the memory corresponds to which original position in the input sequence, both (i) to perform the update, and (ii) to compute the correct attention mask and relative positions.

\subsection{Properties}

\paragraph{Recursive composition.} Transformer Grammars accomplish recursive compositions via a custom attention mask that reflects the hierarchical phrase structures within natural language. Although the mask at a position $i+1$ depends on the mask at position $i$, during training the entire attention mask matrix can be precomputed in advance, and then applied independently to compute multiple syntactic compositions in parallel for the whole segment.
For instance, in the example sequence from Figure~\ref{fig:processing}, during training the representations of NP and VP are computed in parallel, even though their closing nonterminals are at different positions ($6$ and $10$, respectively) in the sequence. Every following layer of Transformer Grammars then takes the composed representations at previous layers, and composes them further. For instance, at position $12$, the second layer will form a composed representation of a sentence constituent \texttt{S)} by using as input the first layer representations of \texttt{NP)} and \texttt{VP)}.
A consequence of this approach is that at least $d$ layers are needed for tokens of depth $d$ to affect the topmost composed representation, a property it shares with conventional Transformers applied to trees~\citep{hahn:2020}.

\paragraph{Context-modulated composition.} TGs' composition steps use a \textsc{compose} attention mask at each closing nonterminal of type \texttt{CNT1}, and all other actions use a \textsc{stack} attention mask. The stack mask makes available the representations of the completed constituents, words, and open nonterminals on the stack. Thus, in the example in Figure~\ref{fig:processing}, the word \texttt{sings} can attend to the closed constituent \texttt{NP)}, as well as ancestor nonterminals \texttt{(S} and \texttt{(VP}. But, importantly, at \texttt{sings}, information about all preceding words is accessible only through the composed \texttt{NP)} representation, thus enforcing the \emph{compressive} effect of syntactic composition.

At higher layers, \textsc{stack} and \textsc{compose} attentions have a subtle interaction worth making explicit. The \textsc{stack} attention that is used to compute the representation for \texttt{sings} can look at the composed representation of the preceding subject \texttt{NP)}, meaning that a certain amount of ``outside information'' can enter into the computation of the composed VP. The availability of outside information deviates from the strict bottom-up compositionality of RNNGs and similar models.

How does this outside information impact composition? In TGs (in contrast to Transformer-XLs), the influence of outside context on composed representations is indirect, and we therefore argue that the TG learner has a bias against capturing such outside information in the composed representation. Our argument relies on two facts: (i)~that learning to compose a representation of a constituent ending at position $i$ is driven by predictions/prediction failures of a subsequent symbol $a_j$, where $j>i$ and (ii)~that if $a_j$'s prediction \emph{does} crucially depend information outside of the constituent ending at $i$, then there will \emph{always} be a more direct attention path than the one via the composed representation at $a_i$. The existence of two paths with different numbers of operations---a more direct one (directly via attention) and a less direct one (via composition followed by attention)---explains the bias against including outside information in composed representations, and in favor of bottom-up information.

Finally, we remark that questions of whether and how contextual information plays a role in composition are complex and unresolved. \citet{bowman-etal-2016-fast} showed that allowing outside information to modulate compositional computations leads to better composed representations, and justified this design on the grounds that outside information may play a crucial disambiguating role in the composition function.

\section{Experiments}

We compare Transformer Grammars with two Transformer-XL baselines: (i) one trained only on the terminal word sequences (\textbf{TXL (terminals)}), and (ii) another trained on the linearized tree sequence as done by \citet{choe-charniak-2016-parsing}, henceforth denoted as \textbf{TXL (trees)}. We remark that model (i) is a word-level language model that estimates the probability of surface strings $p(\boldsymbol{x})$, whereas model (ii) is a syntactic language model that estimates $p(\boldsymbol{x,y})$. We additionally compare against two prior syntactic LMs: (i) the ``generative parsing as language modeling'' approach of \citet{qian-etal-2021-structural}, which operates in a similar fashion as the linearized \citet{choe-charniak-2016-parsing} baseline, albeit with two attention heads that are specialized for syntax (though differently from TGs' explicit recursive syntactic compositions); and (ii) the batched RNNG model of \citet{noji-oseki-2021-effective}.

\paragraph{Datasets.} We conduct experiments on both the Penn Treebank \citep[PTB]{marcus93} dataset ($\approx1$M words), and the BLLIP\textsc{-lg} \citep{charniak2000bllip} dataset according to the split by \citet{Hu:et-al:2020} ($\approx40$M words).
We use the parsed, pre-processed, sentence-level PTB dataset of \citet{dyer-etal-2016-recurrent}, where unseen words and singletons on the training set are mapped according to a special set of unknown word symbols as proposed by \citet{petrov07parser}.
For BLLIP\textsc{-lg}, we use the parse trees provided by \citet{Hu:et-al:2020}. Tokenization is performed with SentencePiece~\cite{kudo-richardson-2018-sentencepiece} using a unigram language model subword algorithm~\cite{kudo-2018-subword} and a vocabulary of 32K word-pieces.
For BLLIP\textsc{-lg}, we consider two settings: (i) we model each sentence independently and (ii) we model each document---each of which is composed of multiple sentences---independently.

\paragraph{Experimental details.} To account for training variance, for each model (TGs, TXL (terminals), and TXL (trees)), we train 100 models of the same size with independent random initializations. On PTB, we use 16-layer models with 12M parameters; whereas on BLLIP\textsc{-lg}, we use 16-layer models with 252M parameters. We select for each training run the model checkpoint with the lowest validation loss, computed using with a single gold proposal tree for each sentence.

\subsection{Language modeling perplexity}

\paragraph{Experimental setup} 

Whereas the probability of a string $\boldsymbol{x}$ can be computed directly by left-to-right decomposition for models operating on strings, for models operating on the joint distribution of strings and syntax trees, we \emph{define} $p(\boldsymbol{x})$ as the marginal distribution: $p(\boldsymbol{x}) = \sum_{\boldsymbol{y} \in \mathcal{Y}_{\boldsymbol{x}}} p(\boldsymbol{x}, \boldsymbol{y})$ where $\mathcal{Y}_{\boldsymbol{x}}$ is the set of possible trees for $\boldsymbol{x}$. As the cardinality of this set is infinite, exact computation of this probability is intractable. However, we can compute a lower bound on $p(\boldsymbol{x})$ by approximately marginalizing over a much smaller set of proposal trees $\mathcal{Y'}_{\boldsymbol{x}}=\{\boldsymbol{y}^{(1)}, \cdots,\boldsymbol{y}^{(N)}\} \subset \mathcal{Y}_{\boldsymbol{x}}$.

For a given $\boldsymbol{x}$, we would want $\mathcal{Y'}_{\boldsymbol{x}}$ to be the set of trees for which $p(\boldsymbol{y} \mid \boldsymbol{x})$ is largest. As this parsing distribution is unavailable, we approximate it with a proposal model $q(\boldsymbol{y} \mid \boldsymbol{x})$. The better this approximation is, the tighter the upper lower bound on $p(\boldsymbol{x})$ is. We use as $q(\boldsymbol{y} \mid \boldsymbol{x})$ a separately-trained discriminative RNNG, and $\mathcal{Y'}_{\boldsymbol{x}}$ is a set of $N = 300$ trees, sampled \emph{without replacement}, as an approximation to the set of $N$ trees with largest $p(\boldsymbol{y} \mid \boldsymbol{x})$. Naturally, regardless of how $\mathcal{Y'}_{\boldsymbol{x}}$ is chosen, the approximate marginal $\sum_{\boldsymbol{y} \in \mathcal{Y'}_{\boldsymbol{x}}} p(\boldsymbol{x}, \boldsymbol{y})$ computed on a subset of $\mathcal{Y}_{\boldsymbol{x}}$ is a lower bound of the true probability $p(\boldsymbol{x})$.

We compute the word perplexity of the validation and test splits of the datasets under the models as $\mathrm{PPL}(\mathcal{D}) = \left( \prod_{\boldsymbol{x} \in \mathcal{D}} p(\boldsymbol{x}) \right)^{-\frac{1}{N_w}}$, where $N_w$ is the total number of words in the dataset $D$. It is exact for the models operating on words, and a conservative approximation (an upper bound) for the models operating on the joint distribution of strings and syntax trees.

For the document-level language models, given a document that consists of $N_s$ sentences, for each sentence $i$ in the document, we need to marginalize over all possible syntax trees for every single $i - 1$ \emph{preceding} sentence in that document. We approximate this by greedily picking the single most likely syntax tree under the model for the first $i - 1$ sentences, before concatenating this single-path prefix with the $300$ tree proposals for the last sentence.

\paragraph{Discussion} 

\begin{figure}[t!]
\centering
\includegraphics{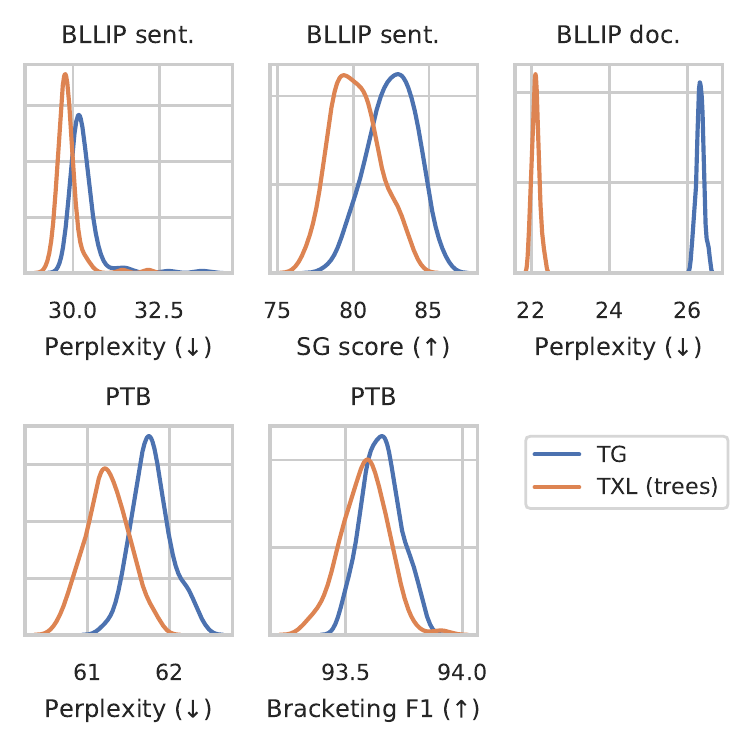}%\vspace{-.3cm}
\caption{Distributions of the metrics of interest on the test sets, with $100$ random initializations for each model. All the differences in means are statistically significant ($p < 10^{-3}$).}
\label{test_distributions}
\end{figure}

We report the mean and sample standard deviation of perplexity (first 3 columns) in Table~\ref{table:main_results}, and plot their distributions in Figure~\ref{test_distributions}.

\begin{table*}[htpb]
\begin{center}
\scalebox{0.9}{
\begin{tabular}{l|ccc|c|c}

{} & \multicolumn{3}{c|}{Perplexity ($\downarrow$)} &            SG ($\uparrow$) &            $F_1$ ($\uparrow$) \\
{} &           PTB &   BLLIP sent. &    BLLIP doc. &   BLLIP sent. &           PTB \\
\hline
TG$^{\dagger}$              &  61.8 $\pm$ 0.2 &  30.3 $\pm$ 0.5 &  26.3 $\pm$ 0.1 &  \textbf{82.5 $\pm$ 1.6} &  \textbf{93.7 $\pm$ 0.1} \\
TXL (trees)$^{\dagger}$        &  \textbf{61.2 $\pm$ 0.3} &  \textbf{29.8 $\pm$ 0.4} &  \textbf{22.1 $\pm$ 0.1} &  80.2 $\pm$ 1.6 &  93.6 $\pm$ 0.1 \\
TXL (terminals) &  62.6 $\pm$ 0.2 &  31.2 $\pm$ 0.4 &  23.1 $\pm$ 0.1 &  69.5 $\pm$ 2.1 &   n/a \\ \midrule
RNNG$^{\diamondsuit}$ \citep{dyer-etal-2016-recurrent} & 105.2 & n/a & n/a & n/a & 93.3 \\
PLM-Mask$^{\diamondsuit}$ \citep{qian-etal-2021-structural} & n/a & 49.1$^{\clubsuit,\dagger}$ $\pm$ 0.3 & n/a & 74.8 & n/a \\
Batched RNNG$^{\diamondsuit}$ \citep{noji-oseki-2021-effective} & n/a & 62.9$^{\heartsuit,\dagger}$  & n/a & 81.4$^{\varheartsuit}$ $\pm$ 2.7 & n/a \\ \midrule
GPT-2 \citep{radford_2019} & n/a & n/a & n/a & 78.4$^{\spadesuit}$ & n/a \\
Gopher \citep{rae_2021}  & n/a & n/a & n/a & 79.5 & n/a\\
Chinchilla \citep{chinchilla}  & n/a & n/a & n/a & 79.7 & n/a
\end{tabular}
}
\end{center}
\caption{Results on the \textbf{test sets} obtained for 100 models of TG, TXL (trees) and TXL.
The results marked $^{\diamondsuit}$ are directly taken from prior work, which may not directly comparable due to differences in model sizes, inference procedures, etc.
% For our results (top three rows), we report the mean and sample standard deviation of the perplexity, bracketing $F_1$ score, and syntactic generalization (SG) score obtained for 100 models of each of the three model variants.
$^{\dagger}$ are upper bounds of perplexities,
% , derived from approximately marginalizing over a set of proposal trees. TXL (terminals) cannot be used to compute bracketing $F_1$ scores. The syntactic generalization test suite assumes models trained on independent sentences from BLLIP\textsc{-lg}. TXL (CC) has the best perplexity overall, whereas TG does best on the tasks that are most related to syntactic structures: SG and bracketing $F_1$ scores.
% In the last two rows (marked with $^{\diamondsuit}$), we report results that are taken directly from prior work.
% We include these results for completeness, although we remark that they are not directly comparable because prior work may use different model sizes, tokenization scheme, etc., although the training dataset for each task (\emph{i.e.,} each column) is exactly identical for all models.
$^{\clubsuit}$ are from personal correspondence with \citeauthor{qian-etal-2021-structural}, and $^{\spadesuit}$ was computed from results by \citet{Hu:et-al:2020}. Results with $^{\heartsuit}$ and $^{\varheartsuit}$ are obtained by personal correspondence with \citeauthor{noji-oseki-2021-effective}; $^{\heartsuit}$we select the best perplexity results for the batched RNNG (35M parameters, beam size of 1,000), $^{\varheartsuit}$model trained on 100M Wikipedia tokens. The PTB parsing results of the batched RNNG \citep[Table 2]{noji-oseki-2021-effective} are not directly comparable since they reported results with beam search inference, whereas we use a parse reranking setup. Perplexities of the large LMs (last 3 rows) are not reported because of likely test data contamination \citep{brown_2020}.
% , whereas result with $^{\spadesuit}$ is approximately derived from a chart \citet[Figure~3]{qian-etal-2021-structural}.
}
\label{table:main_results}
\end{table*}

Although all three models share the exact same number of model parameters and training dataset, our very first observation is that \emph{both} the Transformer-XL baseline trained on the linearized tree sequences (TXL (trees)) and the proposed TG model achieve a lower perplexity---even though the reported perplexity is in fact only an upper bound---than a Transformer-XL trained only on terminals (TXL (terminals))---for which the perplexity calculation is exact---on PTB and the sentence-level BLLIP. This shows that joint modeling of syntactic structures and surface strings in Transformers---even \emph{without} any explicit inductive bias for making use of the syntactic information (\emph{e.g.,} TXL (trees)), is still helpful for improving perplexity.
We conjecture that the next-token prediction task is made easier by the presence of nonterminals within the context, which restricts the word classes that may appear next. Although there are more such prediction events for the linearized tree sequences than for the words-only model, the predictions of the nonterminals are marginalized out at evaluation time. At training time, it might seem that the learning demands placed on the model are higher, and that having to predict the syntactic structures could produce an interference and reduce the available model capacity for predicting the words. Here we do not find this to be the case, as evidenced by both syntactic language models' better perplexity.

Comparing TGs to TXL (trees), perplexity suffers by about 0.5 points (1\%-1.7\% increase relative to the TXL (trees) model's perplexity) on the two sentence-level datasets, and by about 4 points (19\% relative increase) at the document-level on BLLIP-\textsc{lg}. We investigate the causes of this degredation in the Analysis section below~(\S\ref{sec:analysis}).

\subsection{Parse reranking}

% \paragraph{Experimental setup} 

As human-annotated syntax trees are available for the PTB test split,
we also compare models in their parsing accuracy by reranking the 300 candidate samples produced by RNNG for each sentence.

% \paragraph{Discussion} 

We report the mean and sample standard deviation of bracketing $F_1$ as computed with EVALB~\citep{evalb} in Table~\ref{table:main_results}, and plot its distribution in Figure~\ref{test_distributions}. We observe that TG does slightly better ($+0.1\%$) than TXL (trees) on this task; the small difference in mean is nevertheless statistically significant (two-sided Welch's unequal variances {\em t}-test, $p < 10^{-3}$). % This shows that TG is a slightly better than TXL (CC) at scoring parse trees, which may be explained by its restricted attention and by its use of composed syntactic representations.

\subsection{Syntactic generalization}

\paragraph{Experimental setup} 

\citet{Hu:et-al:2020} developed a series of test suites that probe the syntactic ability of language models on a large set of syntactic phenomena. The aim of this task is to comprehensively assess the ability of language models to syntactically generalize in a human-like fashion, which constitutes a key feature of human linguistic ability. A model succeeds on a given test case when the probabilities it assigns to specifically crafted examples obey an inequality (or conjunctions thereof) justified by how humans process language.
We use models trained on independent sentences from BLLIP\textsc{-lg}, evaluate on parse trees provided by \citet{Hu:et-al:2020} and generated using an RNNG proposal model, and we report the average syntactic generalization (\textbf{SG}) score across the same set of 31 test suites.

\paragraph{Discussion} 

We report the mean and standard deviation of the average SG score in Table~\ref{table:main_results}, and plot its distribution in Figure~\ref{test_distributions}. 
% Based on these findings, we make two main observations.
Our first observation is that the average SG score is substantially higher for models trained on linearized trees than on words alone---both TG and TXL (trees) outperform TXL (terminals). Interestingly, this also extends to models that are orders of magnitude larger and trained on much more data, such as GPT-2, Gopher~\citep[280B params.]{rae_2021}, and Chinchilla~\citep[70B params]{chinchilla}.
% ---whether they are TXL (CC) or Transformer Grammars---compared to a Transformer-XL baseline trained on words only.
We believe that this result can be explained in three steps. First, the modeling of the structure via the nonterminals by TG and TXL (trees) can be seen, during training, as providing additional syntactic supervision. This enables them to pick, from a large number of candidate trees, good parses for a sentence. Second, as the SG score is computed from inequalities involving model surprisals on \emph{words}, we perform an approximate marginalization step for TG and TXL (trees). In this approximate marginalization, valid parses are therefore heavily weighted. Finally, when the model has a strong preference for syntactically correct parses, the tasks from the test suite become easier, accounting for these models' higher scores. The results of the large LMs show model scale alone is insufficient to offset this effect.

Our second observation is that---comparing Transformer Grammars to TXL (trees)---our approach is most beneficial on tasks that are most related to modeling structure, \emph{i.e.,} parse reranking, in addition to the comprehensive SG test suite. On both tasks, Transformer Grammars achieve higher bracketing $F_1$ and average SG scores with a statistically significant difference.
% (two-sided Welch's unequal variances $t$-test, difference in means is statistically significant with $p < 10^{-3}$).
We believe that this performance is explained by the restricted attention in Transformer Grammars, thus preventing the model from attending to syntactically irrelevant parts of the input, and encouraging it to learn informative composed representations of subtrees.

\begin{figure}[t]
\centering
\includegraphics{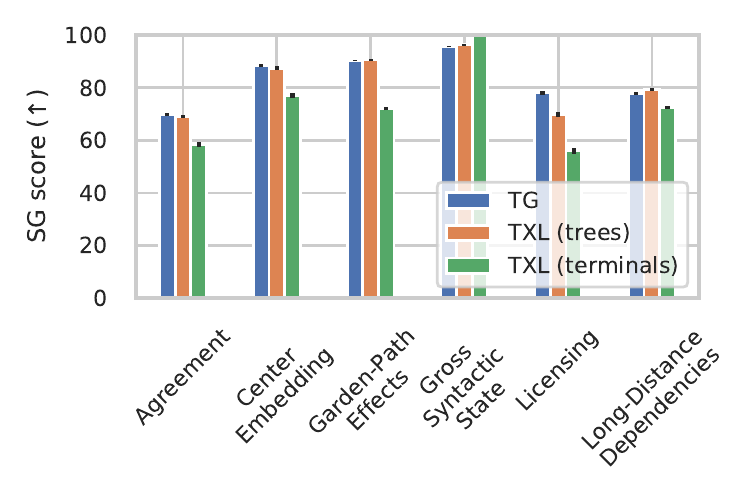}%\vspace{-.5cm}
\caption{Per-circuit breakdown of the SG scores.%, as split by the six circuits.
}
\label{sg_circuits}
\end{figure}

In Figure~\ref{sg_circuits}, we present a breakdown of the SG results. As expected from the average SG score, the TXL (terminals) performs worse than both the TXL (trees) and TG, except for Gross Syntactic State where it nearly reaches 100\%. TG and TXL (trees) have very similar scores on all circuits except licensing, where TG substantially outperforms TXL (trees).\footnote{One might wonder why Licensing and Gross syntactic state deviate from the pattern of results seen in other circuits. Licensing, where TGs excel, involves evaluating restrictions on pairs of words that are linearly separated, but ``structurally local'' (specifically, they stand in a c-command relationship). Since TGs are strongly biased to learn to make predictions in terms of such structurally local relations, it is unsurprising to find success in Licensing. On the other hand, success in Gross Syntactic State requires tracking whether an initial clause is subordinate (in which case a main clause should follow), or a main clause (in which case the sentence should end). Since subordinate clauses are introduced by one of a few subordinating conjunctions, and the content of a main clause is relatively independent of its subordinate, a syntax-free learner will easily learn that the subordinating conjunction determines the Gross syntactic state, explaining the result.} Altogether, these results demonstrate the benefits of recursive syntactic compositions for improving LM performance at syntax-sensitive benchmarks of human linguistic competence, even in the case of powerful Transformer-based language models that are trained at the medium data scale ($\approx40$M words). Furthermore, our findings shed more light on which syntactic constructions benefit the most from explicit syntactic compositions.

\section{Analysis}
\label{sec:analysis}

To better understand what is causing the pattern of results in the previous section, we perform a series of analysis experiments: ablations, comparative analysis of emitted probabilities, and probing of the representations.
\subsection{Ablation experiments}

\begin{table*}[htpb]
\begin{center}

\scalebox{0.9}{
\begin{tabular}{l|ccc|c|c}
{} & \multicolumn{3}{c|}{Perplexity ($\downarrow$)} &          SG ($\uparrow$) &          $F_1$ ($\uparrow$) \\
{} &           PTB & BLLIP sent. &  BLLIP doc. & BLLIP sent. &         PTB \\
\hline
TG$^{\dagger}$                        &    76.1 $\pm$ 0.3 &  31.5 $\pm$ 0.6 &  26.9 $\pm$ 0.1 &  82.5 $\pm$ 1.6 &  \textbf{92.4 $\pm$ 0.1} \\
TG (no position info.)$^{\dagger}$              &    77.5 $\pm$ 0.3 &  31.9 $\pm$ 0.4 &  27.2 $\pm$ 0.1 &  \textbf{82.8 $\pm$ 1.4} &  92.3 $\pm$ 0.1 \\
TG (diff. in linear positions) &    76.3 $\pm$ 0.3 &  31.5 $\pm$ 0.6 &  26.2 $\pm$ 0.1 &  82.6 $\pm$ 1.8 &  \textbf{92.4 $\pm$ 0.1} \\
TG (left-branching)$^{\dagger}$       &   134.0 $\pm$ 0.8 &  45.4 $\pm$ 0.6 &  41.3 $\pm$ 0.2 &  57.2 $\pm$ 2.2 &   n/a \\
TG (right-branching)$^{\dagger}$      &    91.2 $\pm$ 0.4 &  34.2 $\pm$ 0.6 &  29.2 $\pm$ 0.1 &  52.1 $\pm$ 3.8 &   n/a \\
TG (reversed trees)$^{\dagger}$       &  239.7 $\pm$ 11.5 &  49.3 $\pm$ 2.5 &  43.7 $\pm$ 0.6 &  53.9 $\pm$ 3.2 &  82.0 $\pm$ 0.2 \\
\hline
TXL (trees)$^{\dagger}$                  &    \textbf{74.9 $\pm$ 0.3} &  \textbf{30.9 $\pm$ 0.4} &  \textbf{22.3 $\pm$ 0.1} &  80.2 $\pm$ 1.6 &  92.3 $\pm$ 0.1 \\
TXL (trees, no position info.)$^{\dagger}$         &    80.4 $\pm$ 0.3 &  31.9 $\pm$ 0.3 &  31.5 $\pm$ 0.6 &  80.2 $\pm$ 1.8 &  92.0 $\pm$ 0.1 \\
TXL (trees, left-branching)$^{\dagger}$  &   101.0 $\pm$ 0.9 &  34.3 $\pm$ 0.2 &  25.2 $\pm$ 0.2 &  64.3 $\pm$ 3.2 &   n/a \\
TXL (trees, right-branching)$^{\dagger}$ &    97.4 $\pm$ 0.4 &  33.6 $\pm$ 0.3 &  24.4 $\pm$ 0.1 &  53.4 $\pm$ 3.6 &   n/a \\
TXL (trees, reversed trees)$^{\dagger}$  &   129.3 $\pm$ 1.4 &  38.5 $\pm$ 2.6 &  28.3 $\pm$ 0.2 &  57.3 $\pm$ 3.4 &  87.3 $\pm$ 0.2 \\
\hline
TXL (terminals)           &    77.3 $\pm$ 0.3 &  32.2 $\pm$ 0.5 &  23.3 $\pm$ 0.1 &  69.5 $\pm$ 2.1 &   n/a \\
TXL (terminals, no position info.)  &    80.5 $\pm$ 0.3 &  32.9 $\pm$ 0.2 &  25.7 $\pm$ 0.2 &  68.7 $\pm$ 2.0 &   n/a \\
\end{tabular}
}

% \end{tabular}
\end{center}
\caption{Results on the \textbf{validation} split of the datasets. $^{\dagger}$Perplexities reported for TG (all variants) and TXL (trees) are upper bounds, derived from approximately marginalizing over a set of proposal trees.}

\label{table:validation_results}
\end{table*}

As TGs jointly model the words and syntax tree, we perform ablations experiments where the words are preserved, but where the syntactic structure is transformed. The aim of this experiment is to better understand to what extent is having access to the ``right'' syntactic structures at training time an important factor behind TGs' success? Would TGs still do just as well when they are trained with trivial or deterministically transformed syntax trees? We also study the effect of different kinds of position information on our metrics.

\subsubsection{Transformed structures}

We transform a syntax tree into a \emph{binary left-branching} one by moving the opening nonterminals to the left without reordering them, and, after the first two terminals, placing the required closing nonterminal after each terminal. For instance, \texttt{(S (NP the blue bird NP) (VP sings VP) S)} is transformed into \texttt{(S (NP (VP the blue VP) bird NP) sings S)}. Symmetrically, we form a \emph{binary right-branching} tree from this example, \texttt{(S the (NP blue (VP bird sings VP) NP) S)}. Lastly, we define the \emph{reversed trees} where the structure is reversed, \emph{i.e.,} the children of a node are put in reverse order, but the order of the terminals is preserved, in this case \texttt{(S (VP the VP) (NP blue bird sings NP) S)}. In all three transformations, the apparent syntactic structure, as indicated by the nonterminals, is no longer the true syntactic structure of the sentence.

We train and evaluate (on the validation sets) TG and TXL (trees) on such transformed trees, and report our results in Table~\ref{table:validation_results}. For TG, perplexity, syntactic generalization, and bracketing $F_1$\footnote{Here, we do not use the \texttt{DELETE\_LABEL} directives from \texttt{COLLINS.prm}.} are much worse, regardless of the transformation, compared to using the original trees. This is unsurprising considering that the operations performed by the model mechanically depend on the syntactic structure represented by the nonterminals. An apparent structure that does not correspond to the sentence will therefore lead to an unsuitable sequence of operations. More precisely, perplexity is most impacted on reversed trees than on left-branching trees, and right-branching trees have the least impact. Indeed, left-branching trees are comparatively easier to model, because these sequences are formed of a prefix of opening nonterminals, followed by an interleaved sequence of terminals and closing nonterminals. Right-branching trees are similarly easy, and furthermore, the \textsc{compose} operations specific to TG only happen when the closing nonterminals are encountered towards the end of the sequence, which is deterministically determined by its left context.

Unlike TG, TXLs (trees) have no constraints to use the syntactic information to model terminals, and thus they are free to use it or not. However, it is training data, and model capacity must be used to account for its distribution. This explains why performance degradation the TXL (trees) performance is harmed by the tree transformations, but less than the TG performance.

\subsubsection{Positional information}

As observed by \citet{havivTransformerLanguageModels2022}, not using positional information for TXL (terminals) and TXL (trees) has a negative but small impact on perplexity, which the authors conjecture is due to the ability of the model to learn positional information using the causal mask. Under this hypothesis, it follows that the impact on the syntactic generalization and the bracketing $F_1$ scores should be small, which is what we observe. The most impacted model is TXL (trees) on BLLIP-\textsc{lg} documents, which we conjecture is due to the long sequences of tokens, including repeated identical nonterminals, making access to a good positional signal important. For TG, the results are similar, and the same mechanism can be posited to be at play. Its attention mask is not only causal, but also very sparse. Because there are few tokens that can be attended to, position-based querying is at the same time less critical and easier to learn.

We train and evaluate a variant of TG using difference in linear position as relative position function, instead of difference in tree depth, and find almost no impact on performance. This is readily explained by the same reasons as above---as TG's attention mask is so sparse, position-based querying matters little. Given the same empirical performance, we solely ground our choice in its theoretical justification (see \S\ref{sec:relpos}).

\subsection{Regression analysis of probabilities}
\label{analysis_of_log_probs}

To determine when TGs are more or less successful than the unrestricted TXL (trees) model, we predict the differences in log probabilities of the true terminal $a_i$ under the two models: $$\Delta_i = \log p_{\mathrm{TG}}(a_i \mid \boldsymbol{a}_{< i}) - \log p_{\mathrm{TXL (trees)}}(a_i \mid \boldsymbol{a}_{< i}).$$ To reduce variance stemming from model initialization, we use an ensemble of 100 Transformer Grammars and an ensemble of 100 TXLs (trees).

\subsubsection{Terminal frequencies}

We hypothesize that the syntactically-restricted attention pattern of TGs---where subsequent predictions can only attend to composed representations---prevents it from learning the non-syntactic dimensions of the data distribution, such as rare co-occurrences, to the same extent as TXLs (trees). Based on this hypothesis, we expect the TGs' predictions to be worse for rare tokens. 

We therefore compute the empirical unigram distribution of the terminals in the training split of BLLIP-\textsc{lg} documents, and partition terminals into high-frequency ($f \ge 10^{-3}$), medium-frequency ($10^{-5} \le f < 10^{-3}$), and low-frequency ($f < 10^{-5}$) buckets. We then define three binary variables, indicating whether the terminal at a given position has a high, medium, or low frequency, and use these in an ordinary least squares model to predict the difference in log probabilities on the BLLIP-\textsc{lg} validation set: $\Delta \sim \mathrm{HighFreq} + \mathrm{MediumFreq} + \mathrm{LowFreq}$. 

We find an adjusted $R^2$ value of $0.039$, and coefficients $\beta_\mathrm{HighFreq} = -0.0488$, $\beta_\mathrm{MediumFreq} = -0.2419$, $\beta_\mathrm{LowFreq} = -0.5481$, all statistically different from 0 with a $p$-value $< 10^{-3}$.

This shows that---although TGs can predict the terminals appearing most frequently almost as well as TXLs (trees) do---they struggle to predict rarer ones.  We hypothesize that lexical co-occurrences that cross syntactic units can be learnt directly by TXLs (trees), whereas this is more difficult to do for TGs. Indeed, a consequence of \textsc{stack}/\textsc{compose} attention is that a terminal A can only attend to another terminal B \emph{iff} B is in A's left-context, and B is A's sibling. Our result suggests that this is not happening sufficiently often for TGs to predict rare terminals as well as TXLs (trees) do.

\subsubsection{Copying}

Likewise, we hypothesize that TXL (trees) is better at copying words from the context than are TGs.

We define three binary variables, indicating (i)~whether the true terminal to predict appears in the context in a previous sentence, but not in the current one; (ii)~whether it appears in the context in the current sentence; or (iii)~does not appear in the context at all. We use these in a new ordinary least squares model to predict the difference in log probabilities: $\Delta \sim \mathrm{InContextPrevSentences} + \mathrm{InContextCurSentence} + \mathrm{NotInContext}$. 

We find an adjusted $R^2$ value of $0.010$, and coefficients $\beta_\mathrm{InContextPrevSentences} = -0.2871$, $\beta_\mathrm{InContextCurSentence} = -0.1003$, $\beta_\mathrm{NotInContext} = -0.1340$, all statistically different from 0 with a $p$-value $< 10^{-3}$.
This finding suggests that TGs perform worse than TXL (trees) on all three conditions, although the difference is most pronounced for terminals appearing in a previous sentence (but not in the current one). This observation suggests that TXLs (trees) benefit from a priming effect---previously seen tokens becoming more likely---whereas this effect is diminished in TGs.

\subsection{Probing analysis of representations}

\paragraph{Experimental setup}

To quantify how well the representations learnt by TG, TXL (trees) and TXL (terminals) encode syntactic information, we use the information-theoretic probing framework developed by~\citet{voita-titov-2020-information}, capturing in a principled way how well the probes explain the labels given the representations, as well as how readily available the information is in the representations (\emph{i.e.}, how complex the probes are). If the probe labels can be easily predicted from the model's hidden state activations, even with a simple probe, then this provides an indication that the probed phenomenon is more saliently encoded within the model's learnt vector representations. Here we used four probing tasks. The first two probing tasks are taken from \citet{liu2019linguistic}, where we predict (i) the part-of-speech tag of each terminal, and (ii) the grandparent constituent tag of each completed phrase/subtree, which requires an understanding of the relevant phrase-structure information. We then use two other probing tasks from \citet{conneau2018you}: (iii) Predicting the top constituent of the syntax tree and (iv) word-content tagging. We train the probes on the activations output by each layer on the training set of BLLIP-\textsc{lg}, then evaluate the code length on the validation set. For the POS tagging task, we use a single representation for each word. For the three other tasks, we use the representation corresponding to the closing nonterminal for each subtree---for TG, two such representations are available due to the closing nonterminal duplication (\S\ref{sec:recursive_composition}), and we consider them separately; for TXL (terminals), we aggregate the representations of the first and last words of the phrase/subtree.

\paragraph{Discussion}

\begin{figure*}[htbp]
\centering
\includegraphics{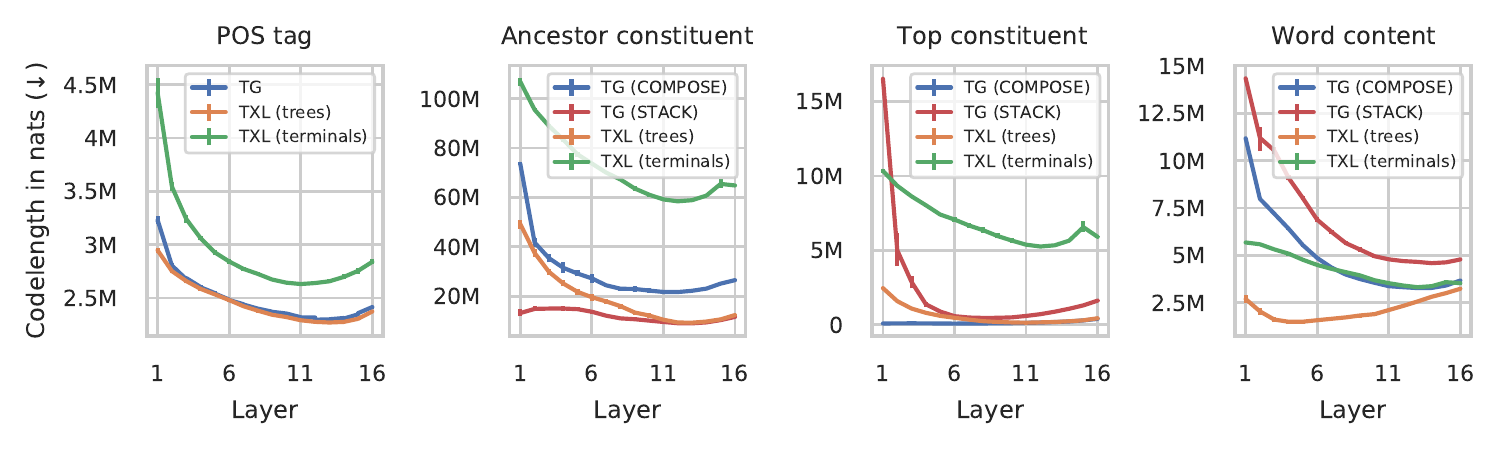}%\vspace{-.5cm}
\caption{Code length of the labels given the model representations on probing tasks.}
\label{probing}
\end{figure*}

We report in Figure \ref{probing} the code length of the probing labels given the model representations on the four probing tasks. We observe that the representations from TG and TXL (trees) explain equally well the POS tags. Predicting the ancestor constituents is much easier using the representations from the closing nonterminals in the \textsc{stack} positions in TG compared to TXL (trees), which is expected considering that the ancestors, by construction of the tree, are still on the stack when the subtree is closed. Conversely, predicting the top constituents (\emph{i.e.}, the sequence of immediate children) is easy to do from the \textsc{compose} representations, which precisely attend over them. On these three syntactic tasks, TXL (terminals), which does not benefit from syntactic supervision, does predictably worse. Finally, it is much easier to predict whether a subtree contains a word from a given set from the representations from TXLs (trees) compared to either type of representations from TG, and even from the TXL (terminals), suggesting that TXL (trees) models retain which words appear in the context better than TGs do, echoing our regression analysis results.

\section{Related Work}

A variety of work has augmented language models with syntactic or hierarchical biases. The RNNG model~\citep{dyer-etal-2016-recurrent,kuncoro-etal-2017-recurrent} jointly models trees and strings and uses recursive networks to build representations of phrases (similar to the approach taken here); however, scaling RNNG training is nontrivial~\citep{noji-oseki-2021-effective}. Other forms of structural bias that do not use observed syntax trees have come in the form of stack-structured memory~\citep{yogatama:2018}, running RNNs at multiple scales~\citep{chung_2016}, and structuring the ``forget'' gates in LSTMs to encourage hierarchy~\citep{shen_2019}.

With the advent of Transformers and large-scale pretraining, the question of whether syntax and hierarchy ``is still needed'' received renewed interest. While bidirectional encoders do learn a great deal about syntax from pretraining~\citep{manning_2020}, long-tail syntactic phenomena continue to pose a problem, and may be implicated in systematic semantic failures~\cite{ettinger-2020-bert}. Prior work has devised multiple strategies for injecting syntactic inductive biases into Transformers~\citep{wang2019structbert,sundararaman2019syntax,kuncoro_2020,sachan-etal-2021-syntax,bai-etal-2021-syntax}. However, improved syntactic awareness is not found to be beneficial for some language understanding tasks \citep{warstadt_2020,kuncoro_2020,pruksachatkun-etal-2020-intermediate,sachan-etal-2021-syntax}.

Our approach to injecting syntactic biases into generative transformer language models combines two modeling traditions: (i) syntactic language models that estimate the joint probability of strings and trees \citep{jurafsky_1995,chelba:2000,roark:2001,henderson:2004,mirowski_2015,choe-charniak-2016-parsing,kim_2019}, and (ii) constraining attention patterns in accordance with syntactic structures \citep{strubell-etal-2018-linguistically,wang-etal-2019-tree,peng-etal-2019-palm,zhang2019sgnet,nguyen_2020,astudillo-etal-2020-transition}. TGs are perhaps most closely related to the model proposed in \citet{qian-etal-2021-structural}, who similarly combined syntactic language modeling with syntax-based attention constraints. We differ from this model in two primary ways. First, TGs use a new kind of typed-attention mask with duplicated closing nonterminal symbols that implement recursive syntactic compositions, which was identified as a critical component of RNN-based syntax models~\citep{dyer-etal-2016-recurrent,kim_2019,wilcox_2019,futrell_2019}. Second, this paper explores an extension of sentence-level syntactic models to models of full documents. Modeling of multi-sentence sequences has been a key feature behind recent language modeling successes~\citep{radford_2019,brown_2020}, and thus understanding how syntax interacts with this modeling problem is of considerable interest.

\section{Conclusion}

Transformer Grammars are a new syntactic language model that implements recursive syntactic composition of phrase representations through attention. Experiments show that TGs outperform prior work on two syntax-sensitive language modeling evaluation metrics. On sentence-level language modeling, TGs outperform a strong Transformer-XL that operates only on the word sequences, although we find that they perform worse at the document-level, when restricted to use a single composed representation for each previous sentence. We also find that the presence of structural information is strictly better on all metrics than in Transformers trained on words alone. While just as efficient to sample from as any autoregressive language model, TGs however do not provide probability estimates as easily, and using it where these are needed requires accepting more computation, or further research into efficient methods for probability estimation. Taken more broadly, our findings emphasize the ongoing importance of finding better and scalable ways to encourage language models to internalize the structural properties of language. Our implementation is available upon request.

\section*{Acknowledgements}

We wish to thank the anonymous reviewers for their feedback and suggestions, as well as the action editors. We would also like to thank Jennifer Hu and Peng Qian for providing us with the BLLIP-\textsc{lg} reparsed data, the partial trees used for the syntactic models evaluated in~\citet{Hu:et-al:2020}, and for their answers to our many questions. We also thank Hiroshi Noji and Yohei Oseki for providing a detailed result breakdown of their efficiently batched RNNG model. Finally, we are grateful to Mark Johnson, Kris Cao, Laura Rimell, Nando de Freitas, and our colleagues in the DeepMind Language team for their insightful thoughts and comments.

\bibliography{custom_subset}

\begin{thebibliography}{55}
\expandafter\ifx\csname natexlab\endcsname\relax\def\natexlab#1{#1}\fi

\bibitem[{Astudillo et~al.(2020)Astudillo, Ballesteros, Naseem, Blodgett, and
  Florian}]{astudillo-etal-2020-transition}
Ram{\'o}n~Fernandez Astudillo, Miguel Ballesteros, Tahira Naseem, Austin
  Blodgett, and Radu Florian. 2020.
\newblock \href {https://doi.org/10.18653/v1/2020.findings-emnlp.89}
  {{Transition-based Parsing with Stack-Transformers}}.
\newblock In \emph{Findings of EMNLP}.

\bibitem[{Bai et~al.(2021)Bai, Wang, Chen, Yang, Bai, Yu, and
  Tong}]{bai-etal-2021-syntax}
Jiangang Bai, Yujing Wang, Yiren Chen, Yaming Yang, Jing Bai, Jing Yu, and
  Yunhai Tong. 2021.
\newblock \href {https://aclanthology.org/2021.eacl-main.262} {{Syntax-{BERT}:
  Improving Pre-trained Transformers with Syntax Trees}}.
\newblock In \emph{Proceedings of the 16th Conference of the European Chapter
  of the Association for Computational Linguistics: Main Volume}, pages
  3011--3020, Online. Association for Computational Linguistics.

\bibitem[{Bowman et~al.(2016)Bowman, Gauthier, Rastogi, Gupta, Manning, and
  Potts}]{bowman-etal-2016-fast}
Samuel~R. Bowman, Jon Gauthier, Abhinav Rastogi, Raghav Gupta, Christopher~D.
  Manning, and Christopher Potts. 2016.
\newblock \href {https://doi.org/10.18653/v1/P16-1139} {{A Fast Unified Model
  for Parsing and Sentence Understanding}}.
\newblock In \emph{Proceedings of the 54th Annual Meeting of the Association
  for Computational Linguistics (Volume 1: Long Papers)}, pages 1466--1477,
  Berlin, Germany. Association for Computational Linguistics.

\bibitem[{Brown et~al.(2020)Brown, Mann, Ryder, Subbiah, Kaplan, Dhariwal,
  Neelakantan, Shyam, Sastry, Askell, Agarwal, Herbert-Voss, Krueger, Henighan,
  Child, Ramesh, Ziegler, Wu, Winter, Hesse, Chen, Sigler, Litwin, Gray, Chess,
  Clark, Berner, McCandlish, Radford, Sutskever, and Amodei}]{brown_2020}
Tom Brown, Benjamin Mann, Nick Ryder, Melanie Subbiah, Jared Kaplan, Prafulla
  Dhariwal, Arvind Neelakantan, Pranav Shyam, Girish Sastry, Amanda Askell,
  Sandhini Agarwal, Ariel Herbert-Voss, Gretchen Krueger, Tom Henighan, Rewon
  Child, Aditya Ramesh, Daniel Ziegler, Jeffrey Wu, Clemens Winter, Chris
  Hesse, Mark Chen, Eric Sigler, Mateusz Litwin, Scott Gray, Benjamin Chess,
  Jack Clark, Christopher Berner, Sam McCandlish, Alec Radford, Ilya Sutskever,
  and Dario Amodei. 2020.
\newblock \href
  {https://proceedings.neurips.cc/paper/2020/file/1457c0d6bfcb4967418bfb8ac142f64a-Paper.pdf}
  {{Language Models are Few-Shot Learners}}.
\newblock In \emph{Advances in Neural Information Processing Systems},
  volume~33, pages 1877--1901. Curran Associates, Inc.

\bibitem[{Charniak et~al.(2000)Charniak, Blaheta, Ge, Hall, Hale, and
  Johnson}]{charniak2000bllip}
Eugene Charniak, Don Blaheta, Niyu Ge, Keith Hall, John Hale, and Mark Johnson.
  2000.
\newblock \href {https://catalog.ldc.upenn.edu/LDC2000T43} {{BLLIP 1987--89 WSJ
  Corpus Release 1, LDC2000T43}}.
\newblock \emph{LDC2000T43. Linguistic Data Consortium}, 36.

\bibitem[{Chelba and Jelinek(2000)}]{chelba:2000}
Ciprian Chelba and Frederick Jelinek. 2000.
\newblock \href {https://doi.org/https://doi.org/10.1006/csla.2000.0147}
  {Structured language modeling}.
\newblock \emph{Computer Speech \& Language}, 14(4):283--332.

\bibitem[{Choe and Charniak(2016)}]{choe-charniak-2016-parsing}
Do~Kook Choe and Eugene Charniak. 2016.
\newblock \href {https://doi.org/10.18653/v1/D16-1257} {{Parsing as Language
  Modeling}}.
\newblock In \emph{Proceedings of the 2016 Conference on Empirical Methods in
  Natural Language Processing}, pages 2331--2336, Austin, Texas. Association
  for Computational Linguistics.

\bibitem[{Chomsky(1957)}]{chomsky_1957}
Noam Chomsky. 1957.
\newblock \href
  {https://www.degruyter.com/document/doi/10.1515/9783112316009/html?lang=en}
  {\emph{{Syntactic Structures}}}.
\newblock Mouton, The Hague/Paris.

\bibitem[{Chung et~al.(2017)Chung, Ahn, and Bengio}]{chung_2016}
Junyoung Chung, Sungjin Ahn, and Yoshua Bengio. 2017.
\newblock \href {https://openreview.net/pdf?id=S1di0sfgl} {{Hierarchical
  Multiscale Recurrent Neural Networks}}.
\newblock In \emph{Proc. of ICLR}.

\bibitem[{Conneau et~al.(2018)Conneau, Kruszewski, Lample, Barrault, and
  Baroni}]{conneau2018you}
Alexis Conneau, German Kruszewski, Guillaume Lample, Lo{\"\i}c Barrault, and
  Marco Baroni. 2018.
\newblock \href {https://doi.org/10.18653/v1/P18-1198} {What you can cram into
  a single {\$}{\&}!{\#}* vector: Probing sentence embeddings for linguistic
  properties}.
\newblock In \emph{Proceedings of the 56th Annual Meeting of the Association
  for Computational Linguistics (Volume 1: Long Papers)}, pages 2126--2136,
  Melbourne, Australia. Association for Computational Linguistics.

\bibitem[{Dai et~al.(2019)Dai, Yang, Yang, Carbonell, Le, and
  Salakhutdinov}]{dai-etal-2019-transformer}
Zihang Dai, Zhilin Yang, Yiming Yang, Jaime Carbonell, Quoc Le, and Ruslan
  Salakhutdinov. 2019.
\newblock \href {https://doi.org/10.18653/v1/P19-1285} {{Transformer-{XL}:
  Attentive Language Models beyond a Fixed-Length Context}}.
\newblock In \emph{Proc. of ACL}.

\bibitem[{Devlin et~al.(2019)Devlin, Chang, Lee, and Toutanova}]{devlin_2019}
Jacob Devlin, Ming-Wei Chang, Kenton Lee, and Kristina Toutanova. 2019.
\newblock \href {https://doi.org/10.18653/v1/N19-1423} {{{BERT}: Pre-training
  of Deep Bidirectional Transformers for Language Understanding}}.
\newblock In \emph{Proc. of NAACL}.

\bibitem[{Dyer et~al.(2016)Dyer, Kuncoro, Ballesteros, and
  Smith}]{dyer-etal-2016-recurrent}
Chris Dyer, Adhiguna Kuncoro, Miguel Ballesteros, and Noah~A. Smith. 2016.
\newblock \href {https://doi.org/10.18653/v1/N16-1024} {{Recurrent Neural
  Network Grammars}}.
\newblock In \emph{Proceedings of the 2016 Conference of the North {A}merican
  Chapter of the Association for Computational Linguistics: Human Language
  Technologies}, pages 199--209, San Diego, California. Association for
  Computational Linguistics.

\bibitem[{Ettinger(2020)}]{ettinger-2020-bert}
Allyson Ettinger. 2020.
\newblock \href {https://doi.org/10.1162/tacl_a_00298} {{What {BERT} Is Not:
  Lessons from a New Suite of Psycholinguistic Diagnostics for Language
  Models}}.
\newblock \emph{Transactions of the Association for Computational Linguistics}.

\bibitem[{Everaert et~al.(2015)Everaert, Huybregts, Chomsky, Berwick, and
  Bolhuis}]{structures:not:strings}
Martin~B.H. Everaert, Marinus~A.C. Huybregts, Noam Chomsky, Robert~C. Berwick,
  and Johan~J. Bolhuis. 2015.
\newblock \href {https://doi.org/https://doi.org/10.1016/j.tics.2015.09.008}
  {{Structures, Not Strings: Linguistics as Part of the Cognitive Sciences}}.
\newblock \emph{Trends in Cognitive Sciences}, 19(12):729--743.

\bibitem[{Futrell et~al.(2019)Futrell, Wilcox, Morita, Qian, Ballesteros, and
  Levy}]{futrell_2019}
Richard Futrell, Ethan Wilcox, Takashi Morita, Peng Qian, Miguel Ballesteros,
  and Roger Levy. 2019.
\newblock \href {https://doi.org/10.18653/v1/N19-1004} {{Neural language models
  as psycholinguistic subjects: Representations of syntactic state}}.
\newblock In \emph{Proceedings of the 2019 Conference of the North {A}merican
  Chapter of the Association for Computational Linguistics: Human Language
  Technologies, Volume 1 (Long and Short Papers)}, pages 32--42, Minneapolis,
  Minnesota. Association for Computational Linguistics.

\bibitem[{Hahn(2020)}]{hahn:2020}
Michael Hahn. 2020.
\newblock \href {https://doi.org/10.1162/tacl_a_00306} {{Theoretical
  Limitations of Self-Attention in Neural Sequence Models}}.
\newblock \emph{Transactions of the Association for Computational Linguistics},
  8:156--171.

\bibitem[{Haviv et~al.(2022)Haviv, Ram, Press, Izsak, and
  Levy}]{havivTransformerLanguageModels2022}
Adi Haviv, Ori Ram, Ofir Press, Peter Izsak, and Omer Levy. 2022.
\newblock \href {http://arxiv.org/abs/2203.16634} {Transformer {{Language
  Models}} without {{Positional Encodings Still Learn Positional
  Information}}}.
\newblock (arXiv:2203.16634).

\bibitem[{Henderson(2004)}]{henderson:2004}
James Henderson. 2004.
\newblock \href {https://doi.org/10.3115/1218955.1218968} {{Discriminative
  Training of a Neural Network Statistical Parser}}.
\newblock In \emph{Proceedings of the 42nd Annual Meeting of the Association
  for Computational Linguistics ({ACL}-04)}, pages 95--102, Barcelona, Spain.

\bibitem[{Hoffmann et~al.(2022)Hoffmann, Borgeaud, Mensch, Buchatskaya, Cai,
  Rutherford, Casas, Hendricks, Welbl, Clark, Hennigan, Noland, Millican,
  Driessche, Damoc, Guy, Osindero, Simonyan, Elsen, Rae, Vinyals, and
  Sifre}]{chinchilla}
Jordan Hoffmann, Sebastian Borgeaud, Arthur Mensch, Elena Buchatskaya, Trevor
  Cai, Eliza Rutherford, Diego de~Las Casas, Lisa~Anne Hendricks, Johannes
  Welbl, Aidan Clark, Tom Hennigan, Eric Noland, Katie Millican, George van~den
  Driessche, Bogdan Damoc, Aurelia Guy, Simon Osindero, Karen Simonyan, Erich
  Elsen, Jack~W. Rae, Oriol Vinyals, and Laurent Sifre. 2022.
\newblock \href {https://doi.org/10.48550/ARXIV.2203.15556} {{Training
  Compute-Optimal Large Language Models}}.
\newblock \emph{CoRR}, abs/2203.15556v1.

\bibitem[{Hu et~al.(2020)Hu, Gauthier, Qian, Wilcox, and Levy}]{Hu:et-al:2020}
Jennifer Hu, Jon Gauthier, Peng Qian, Ethan Wilcox, and Roger Levy. 2020.
\newblock \href {https://doi.org/10.18653/v1/2020.acl-main.158} {{A Systematic
  Assessment of Syntactic Generalization in Neural Language Models}}.
\newblock In \emph{Proceedings of the 58th Annual Meeting of the Association
  for Computational Linguistics}, pages 1725--1744, Online. Association for
  Computational Linguistics.

\bibitem[{Jurafsky et~al.(1995)Jurafsky, Wooters, Segal, Stolcke, Fosler,
  Tajchman, and Morgan}]{jurafsky_1995}
Daniel Jurafsky, Chuck Wooters, Jonathan Segal, Andreas Stolcke, Eric Fosler,
  Gary~N. Tajchman, and Nelson Morgan. 1995.
\newblock \href {https://doi.org/10.1109/ICASSP.1995.479396} {Using a
  stochastic context-free grammar as a language model for speech recognition}.
\newblock In \emph{Proc. of ICASSP}.

\bibitem[{Kim et~al.(2019)Kim, Rush, Yu, Kuncoro, Dyer, and Melis}]{kim_2019}
Yoon Kim, Alexander Rush, Lei Yu, Adhiguna Kuncoro, Chris Dyer, and G{\'a}bor
  Melis. 2019.
\newblock \href {https://doi.org/10.18653/v1/N19-1114} {{Unsupervised Recurrent
  Neural Network Grammars}}.
\newblock In \emph{Proc. of NAACL}.

\bibitem[{Kudo(2018)}]{kudo-2018-subword}
Taku Kudo. 2018.
\newblock \href {https://doi.org/10.18653/v1/P18-1007} {{Subword
  Regularization: Improving Neural Network Translation Models with Multiple
  Subword Candidates}}.
\newblock In \emph{Proceedings of the 56th Annual Meeting of the Association
  for Computational Linguistics (Volume 1: Long Papers)}, pages 66--75,
  Melbourne, Australia. Association for Computational Linguistics.

\bibitem[{Kudo and Richardson(2018)}]{kudo-richardson-2018-sentencepiece}
Taku Kudo and John Richardson. 2018.
\newblock \href {https://doi.org/10.18653/v1/D18-2012} {{{S}entence{P}iece: A
  simple and language independent subword tokenizer and detokenizer for Neural
  Text Processing}}.
\newblock In \emph{Proceedings of the 2018 Conference on Empirical Methods in
  Natural Language Processing: System Demonstrations}, pages 66--71, Brussels,
  Belgium. Association for Computational Linguistics.

\bibitem[{Kuncoro et~al.(2017)Kuncoro, Ballesteros, Kong, Dyer, Neubig, and
  Smith}]{kuncoro-etal-2017-recurrent}
Adhiguna Kuncoro, Miguel Ballesteros, Lingpeng Kong, Chris Dyer, Graham Neubig,
  and Noah~A. Smith. 2017.
\newblock \href {https://aclanthology.org/E17-1117} {{What Do Recurrent Neural
  Network Grammars Learn About Syntax?}}
\newblock In \emph{Proceedings of the 15th Conference of the {E}uropean Chapter
  of the Association for Computational Linguistics: Volume 1, Long Papers},
  pages 1249--1258, Valencia, Spain. Association for Computational Linguistics.

\bibitem[{Kuncoro et~al.(2019)Kuncoro, Dyer, Rimell, Clark, and
  Blunsom}]{kuncoro_19}
Adhiguna Kuncoro, Chris Dyer, Laura Rimell, Stephen Clark, and Phil Blunsom.
  2019.
\newblock \href {https://doi.org/10.18653/v1/P19-1337} {{Scalable Syntax-Aware
  Language Models Using Knowledge Distillation}}.
\newblock In \emph{Proceedings of the 57th Annual Meeting of the Association
  for Computational Linguistics}, pages 3472--3484, Florence, Italy.
  Association for Computational Linguistics.

\bibitem[{Kuncoro et~al.(2020)Kuncoro, Kong, Fried, Yogatama, Rimell, Dyer, and
  Blunsom}]{kuncoro_2020}
Adhiguna Kuncoro, Lingpeng Kong, Daniel Fried, Dani Yogatama, Laura Rimell,
  Chris Dyer, and Phil Blunsom. 2020.
\newblock \href {https://doi.org/10.1162/tacl_a_00345} {{Syntactic Structure
  Distillation Pretraining for Bidirectional Encoders}}.
\newblock \emph{Transactions of the Association for Computational Linguistics},
  8:776--794.

\bibitem[{Liu et~al.(2019)Liu, Gardner, Belinkov, Peters, and
  Smith}]{liu2019linguistic}
Nelson~F. Liu, Matt Gardner, Yonatan Belinkov, Matthew~E. Peters, and Noah~A.
  Smith. 2019.
\newblock \href {https://doi.org/10.18653/v1/N19-1112} {Linguistic knowledge
  and transferability of contextual representations}.
\newblock In \emph{Proceedings of the 2019 Conference of the North {A}merican
  Chapter of the Association for Computational Linguistics: Human Language
  Technologies, Volume 1 (Long and Short Papers)}, pages 1073--1094,
  Minneapolis, Minnesota. Association for Computational Linguistics.

\bibitem[{Manning et~al.(2020)Manning, Clark, Hewitt, Khandelwal, and
  Levy}]{manning_2020}
Christopher~D. Manning, Kevin Clark, John Hewitt, Urvashi Khandelwal, and Omer
  Levy. 2020.
\newblock \href {https://doi.org/10.1073/pnas.1907367117} {Emergent linguistic
  structure in artificial neural networks trained by self-supervision}.
\newblock \emph{Proceedings of the National Academy of Sciences}, 117(48).

\bibitem[{Marcus et~al.(1993)Marcus, Santorini, and Marcinkiewicz}]{marcus93}
Mitchell~P. Marcus, Beatrice Santorini, and Mary~Ann Marcinkiewicz. 1993.
\newblock \href {https://aclanthology.org/J93-2004} {{Building a Large
  Annotated Corpus of {E}nglish: The {P}enn {T}reebank}}.
\newblock \emph{Computational Linguistics}, 19(2):313--330.

\bibitem[{Mirowski and Vlachos(2015)}]{mirowski_2015}
Piotr Mirowski and Andreas Vlachos. 2015.
\newblock \href {https://doi.org/10.3115/v1/P15-2084} {{Dependency Recurrent
  Neural Language Models for Sentence Completion}}.
\newblock In \emph{Proc. of ACL-IJCNLP}.

\bibitem[{Nguyen et~al.(2020)Nguyen, Joty, Hoi, and Socher}]{nguyen_2020}
Xuan{-}Phi Nguyen, Shafiq~R. Joty, Steven C.~H. Hoi, and Richard Socher. 2020.
\newblock \href {https://openreview.net/forum?id=HJxK5pEYvr} {{Tree-Structured
  Attention with Hierarchical Accumulation}}.
\newblock In \emph{8th International Conference on Learning Representations,
  {ICLR} 2020, Addis Ababa, Ethiopia, April 26-30, 2020}. OpenReview.net.

\bibitem[{Noji and Oseki(2021)}]{noji-oseki-2021-effective}
Hiroshi Noji and Yohei Oseki. 2021.
\newblock \href {https://doi.org/10.18653/v1/2021.findings-acl.380} {{Effective
  Batching for Recurrent Neural Network Grammars}}.
\newblock In \emph{Findings of the ACL-IJCNLP}.

\bibitem[{Peng et~al.(2019)Peng, Schwartz, and Smith}]{peng-etal-2019-palm}
Hao Peng, Roy Schwartz, and Noah~A. Smith. 2019.
\newblock \href {https://doi.org/10.18653/v1/D19-1376} {{{P}a{LM}: A Hybrid
  Parser and Language Model}}.
\newblock In \emph{Proc. of EMNLP-IJCNLP}.

\bibitem[{Peters et~al.(2018)Peters, Neumann, Iyyer, Gardner, Clark, Lee, and
  Zettlemoyer}]{peters_2018}
Matthew~E. Peters, Mark Neumann, Mohit Iyyer, Matt Gardner, Christopher Clark,
  Kenton Lee, and Luke Zettlemoyer. 2018.
\newblock \href {https://doi.org/10.18653/v1/N18-1202} {{Deep Contextualized
  Word Representations}}.
\newblock In \emph{Proc. of NAACL}.

\bibitem[{Petrov and Klein(2007)}]{petrov07parser}
Slav Petrov and Dan Klein. 2007.
\newblock \href {https://aclanthology.org/N07-1051} {{Improved Inference for
  Unlexicalized Parsing}}.
\newblock In \emph{Human Language Technologies 2007: The Conference of the
  North {A}merican Chapter of the Association for Computational Linguistics;
  Proceedings of the Main Conference}, pages 404--411, Rochester, New York.
  Association for Computational Linguistics.

\bibitem[{Pruksachatkun et~al.(2020)Pruksachatkun, Phang, Liu, Htut, Zhang,
  Pang, Vania, Kann, and Bowman}]{pruksachatkun-etal-2020-intermediate}
Yada Pruksachatkun, Jason Phang, Haokun Liu, Phu~Mon Htut, Xiaoyi Zhang,
  Richard~Yuanzhe Pang, Clara Vania, Katharina Kann, and Samuel~R. Bowman.
  2020.
\newblock \href {https://doi.org/10.18653/v1/2020.acl-main.467}
  {{Intermediate-Task Transfer Learning with Pretrained Language Models: When
  and Why Does It Work?}}
\newblock In \emph{Proc. of ACL}.

\bibitem[{Qian et~al.(2021)Qian, Naseem, Levy, and
  Fernandez~Astudillo}]{qian-etal-2021-structural}
Peng Qian, Tahira Naseem, Roger Levy, and Ram{\'o}n Fernandez~Astudillo. 2021.
\newblock \href {https://doi.org/10.18653/v1/2021.acl-long.289} {{Structural
  Guidance for Transformer Language Models}}.
\newblock In \emph{Proceedings of the 59th Annual Meeting of the Association
  for Computational Linguistics and the 11th International Joint Conference on
  Natural Language Processing (Volume 1: Long Papers)}, pages 3735--3745,
  Online. Association for Computational Linguistics.

\bibitem[{Radford et~al.(2019)Radford, Wu, Child, Luan, Amodei, and
  Sutskever}]{radford_2019}
Alec Radford, Jeff Wu, Rewon Child, David Luan, Dario Amodei, and Ilya
  Sutskever. 2019.
\newblock \href
  {https://d4mucfpksywv.cloudfront.net/better-language-models/language_models_are_unsupervised_multitask_learners.pdf}
  {{Language Models are Unsupervised Multitask Learners}}.

\bibitem[{Rae et~al.(2021)Rae, Borgeaud, Cai, Millican, Hoffmann, Song,
  Aslanides, Henderson, Ring, Young, Rutherford, Hennigan, Menick, Cassirer,
  Powell, van~den Driessche, Hendricks, Rauh, Huang, Glaese, Welbl, Dathathri,
  Huang, Uesato, Mellor, Higgins, Creswell, McAleese, Wu, Elsen, Jayakumar,
  Buchatskaya, Budden, Sutherland, Simonyan, Paganini, Sifre, Martens, Li,
  Kuncoro, Nematzadeh, Gribovskaya, Donato, Lazaridou, Mensch, Lespiau,
  Tsimpoukelli, Grigorev, Fritz, Sottiaux, Pajarskas, Pohlen, Gong, Toyama,
  de~Masson~d'Autume, Li, Terzi, Mikulik, Babuschkin, Clark, de~Las~Casas, Guy,
  Jones, Bradbury, Johnson, Hechtman, Weidinger, Gabriel, Isaac, Lockhart,
  Osindero, Rimell, Dyer, Vinyals, Ayoub, Stanway, Bennett, Hassabis,
  Kavukcuoglu, and Irving}]{rae_2021}
Jack~W. Rae, Sebastian Borgeaud, Trevor Cai, Katie Millican, Jordan Hoffmann,
  H.~Francis Song, John Aslanides, Sarah Henderson, Roman Ring, Susannah Young,
  Eliza Rutherford, Tom Hennigan, Jacob Menick, Albin Cassirer, Richard Powell,
  George van~den Driessche, Lisa~Anne Hendricks, Maribeth Rauh, Po{-}Sen Huang,
  Amelia Glaese, Johannes Welbl, Sumanth Dathathri, Saffron Huang, Jonathan
  Uesato, John Mellor, Irina Higgins, Antonia Creswell, Nat McAleese, Amy Wu,
  Erich Elsen, Siddhant~M. Jayakumar, Elena Buchatskaya, David Budden, Esme
  Sutherland, Karen Simonyan, Michela Paganini, Laurent Sifre, Lena Martens,
  Xiang~Lorraine Li, Adhiguna Kuncoro, Aida Nematzadeh, Elena Gribovskaya,
  Domenic Donato, Angeliki Lazaridou, Arthur Mensch, Jean{-}Baptiste Lespiau,
  Maria Tsimpoukelli, Nikolai Grigorev, Doug Fritz, Thibault Sottiaux, Mantas
  Pajarskas, Toby Pohlen, Zhitao Gong, Daniel Toyama, Cyprien
  de~Masson~d'Autume, Yujia Li, Tayfun Terzi, Vladimir Mikulik, Igor
  Babuschkin, Aidan Clark, Diego de~Las~Casas, Aurelia Guy, Chris Jones, James
  Bradbury, Matthew Johnson, Blake~A. Hechtman, Laura Weidinger, Iason Gabriel,
  William~S. Isaac, Edward Lockhart, Simon Osindero, Laura Rimell, Chris Dyer,
  Oriol Vinyals, Kareem Ayoub, Jeff Stanway, Lorrayne Bennett, Demis Hassabis,
  Koray Kavukcuoglu, and Geoffrey Irving. 2021.
\newblock \href {https://arxiv.org/abs/2112.11446v2} {{Scaling Language Models:
  Methods, Analysis {\&} Insights from Training Gopher}}.
\newblock \emph{CoRR}, abs/2112.11446v2.

\bibitem[{Roark(2001)}]{roark:2001}
Brian Roark. 2001.
\newblock \href {https://doi.org/10.1162/089120101750300526} {{Probabilistic
  Top-Down Parsing and Language Modeling}}.
\newblock \emph{Computational Linguistics}, 27(2):249--276.

\bibitem[{Sachan et~al.(2021)Sachan, Zhang, Qi, and
  Hamilton}]{sachan-etal-2021-syntax}
Devendra Sachan, Yuhao Zhang, Peng Qi, and William~L. Hamilton. 2021.
\newblock \href {https://doi.org/10.18653/v1/2021.eacl-main.228} {{Do Syntax
  Trees Help Pre-trained Transformers Extract Information?}}
\newblock In \emph{Proc. of EACL}.

\bibitem[{Sekine and Collins(1997)}]{evalb}
Satoshi Sekine and Michael~John Collins. 1997.
\newblock \href {https://nlp.cs.nyu.edu/evalb/} {{Evalb - bracket scoring
  program}}.

\bibitem[{Shen et~al.(2019)Shen, Tan, Sordoni, and Courville}]{shen_2019}
Yikang Shen, Shawn Tan, Alessandro Sordoni, and Aaron Courville. 2019.
\newblock \href {https://openreview.net/forum?id=B1l6qiR5F7} {{Ordered Neurons:
  Integrating Tree Structures into Recurrent Neural Networks}}.
\newblock In \emph{International Conference on Learning Representations}.

\bibitem[{Strubell et~al.(2018)Strubell, Verga, Andor, Weiss, and
  McCallum}]{strubell-etal-2018-linguistically}
Emma Strubell, Patrick Verga, Daniel Andor, David Weiss, and Andrew McCallum.
  2018.
\newblock \href {https://doi.org/10.18653/v1/D18-1548}
  {{Linguistically-Informed Self-Attention for Semantic Role Labeling}}.
\newblock In \emph{Proceedings of the 2018 Conference on Empirical Methods in
  Natural Language Processing}, pages 5027--5038, Brussels, Belgium.
  Association for Computational Linguistics.

\bibitem[{Sundararaman et~al.(2019)Sundararaman, Subramanian, Wang, Si, Shen,
  Wang, and Carin}]{sundararaman2019syntax}
Dhanasekar Sundararaman, Vivek Subramanian, Guoyin Wang, Shijing Si, Dinghan
  Shen, Dong Wang, and Lawrence Carin. 2019.
\newblock \href {https://arxiv.org/abs/1911.06156v1} {{Syntax-Infused
  Transformer and BERT models for Machine Translation and Natural Language
  Understanding}}.
\newblock \emph{arXiv preprint arXiv:1911.06156v1}.

\bibitem[{Vinyals et~al.(2015)Vinyals, Kaiser, Koo, Petrov, Sutskever, and
  Hinton}]{vinyals:2015}
Oriol Vinyals, \L{}ukasz Kaiser, Terry Koo, Slav Petrov, Ilya Sutskever, and
  Geoffrey Hinton. 2015.
\newblock \href
  {https://proceedings.neurips.cc/paper/2015/file/277281aada22045c03945dcb2ca6f2ec-Paper.pdf}
  {{Grammar as a Foreign Language}}.
\newblock In \emph{Advances in Neural Information Processing Systems},
  volume~28. Curran Associates, Inc.

\bibitem[{Voita and Titov(2020)}]{voita-titov-2020-information}
Elena Voita and Ivan Titov. 2020.
\newblock \href {https://doi.org/10.18653/v1/2020.emnlp-main.14}
  {{Information-Theoretic Probing with Minimum Description Length}}.
\newblock In \emph{Proceedings of the 2020 Conference on Empirical Methods in
  Natural Language Processing (EMNLP)}, pages 183--196, Online. Association for
  Computational Linguistics.

\bibitem[{Wang et~al.(2020)Wang, Bi, Yan, Wu, Bao, Peng, and
  Si}]{wang2019structbert}
Wei Wang, Bin Bi, Ming Yan, Chen Wu, Zuyi Bao, Liwei Peng, and Luo Si. 2020.
\newblock \href {https://openreview.net/pdf?id=BJgQ4lSFPH} {{Struct{BERT}:
  Incorporating Language Structures into Pre-training for Deep Language
  Understanding}}.
\newblock In \emph{Proc. of ICLR}.

\bibitem[{Wang et~al.(2019)Wang, Lee, and Chen}]{wang-etal-2019-tree}
Yaushian Wang, Hung-Yi Lee, and Yun-Nung Chen. 2019.
\newblock \href {https://doi.org/10.18653/v1/D19-1098} {{Tree Transformer:
  Integrating Tree Structures into Self-Attention}}.
\newblock In \emph{Proc. of EMNLP-IJCNLP}.

\bibitem[{Warstadt et~al.(2020)Warstadt, Parrish, Liu, Mohananey, Peng, Wang,
  and Bowman}]{warstadt_2020}
Alex Warstadt, Alicia Parrish, Haokun Liu, Anhad Mohananey, Wei Peng, Sheng-Fu
  Wang, and Samuel~R. Bowman. 2020.
\newblock \href {https://doi.org/10.1162/tacl\_a\_00321} {{BLiMP: The Benchmark
  of Linguistic Minimal Pairs for English}}.
\newblock \emph{TACL}.

\bibitem[{Wilcox et~al.(2019)Wilcox, Qian, Futrell, Ballesteros, and
  Levy}]{wilcox_2019}
Ethan Wilcox, Peng Qian, Richard Futrell, Miguel Ballesteros, and Roger Levy.
  2019.
\newblock \href {https://doi.org/10.18653/v1/N19-1334} {{Structural Supervision
  Improves Learning of Non-Local Grammatical Dependencies}}.
\newblock In \emph{Proceedings of the 2019 Conference of the North {A}merican
  Chapter of the Association for Computational Linguistics: Human Language
  Technologies, Volume 1 (Long and Short Papers)}, pages 3302--3312,
  Minneapolis, Minnesota. Association for Computational Linguistics.

\bibitem[{Yogatama et~al.(2018)Yogatama, Miao, Melis, Ling, Kuncoro, Dyer, and
  Blunsom}]{yogatama:2018}
Dani Yogatama, Yishu Miao, Gabor Melis, Wang Ling, Adhiguna Kuncoro, Chris
  Dyer, and Phil Blunsom. 2018.
\newblock \href {https://openreview.net/pdf?id=SkFqf0lAZ} {{Memory
  Architectures in Recurrent Neural Network Language Models}}.
\newblock In \emph{Proc. of ICLR}.

\bibitem[{Zhang et~al.(2020)Zhang, Wu, Zhou, Duan, Zhao, and
  Wang}]{zhang2019sgnet}
Zhuosheng Zhang, Yuwei Wu, Junru Zhou, Sufeng Duan, Hai Zhao, and Rui Wang.
  2020.
\newblock \href {https://ojs.aaai.org/index.php/AAAI/article/view/6511}
  {{SG-Net: Syntax-Guided Machine Reading Comprehension}}.
\newblock In \emph{The Thirty-Fourth {AAAI} Conference on Artificial
  Intelligence, {AAAI} 2020, The Thirty-Second Innovative Applications of
  Artificial Intelligence Conference, {IAAI} 2020, The Tenth {AAAI} Symposium
  on Educational Advances in Artificial Intelligence, {EAAI} 2020, New York,
  NY, USA, February 7-12, 2020}, pages 9636--9643. {AAAI} Press.

\end{thebibliography}
\bibliographystyle{acl_natbib}

\end{document}